\documentclass[conference]{IEEEtran}
\IEEEoverridecommandlockouts
\usepackage{cite}
\usepackage{amsmath,amssymb,amsfonts}
\usepackage{algorithmic}
\usepackage{graphicx}
\usepackage{textcomp}
\usepackage{xcolor}
\usepackage{subcaption}
\usepackage{colortbl}
\usepackage{caption}
\usepackage{cuted}
\usepackage{booktabs}
\usepackage{flushend} 

\usepackage{dblfloatfix}
\def\BibTeX{{\rm B\kern-.05em{\sc i\kern-.025em b}\kern-.08em
    T\kern-.1667em\lower.7ex\hbox{E}\kern-.125emX}}
\begin{document}

\title{Learning Robust Dexterous In-Hand Manipulation from Joint Sensors with Proprioceptive Transformer}

\author{
    \IEEEauthorblockN{
        Senlan Yao\textsuperscript{1},\ 
        Chenyu Yang\textsuperscript{1},\ 
        Jaehoon Kim\textsuperscript{1},\ 
        Aristotelis Sympetheros\textsuperscript{1},\
        Robert K. Katzschmann\textsuperscript{1}
    }
    \IEEEauthorblockA{
        \textsuperscript{1}Soft Robotics Laboratory, ETH Zürich, Switzerland
    }
}

\maketitle
\begin{strip}
    \centering
    \includegraphics[width=0.9\textwidth]{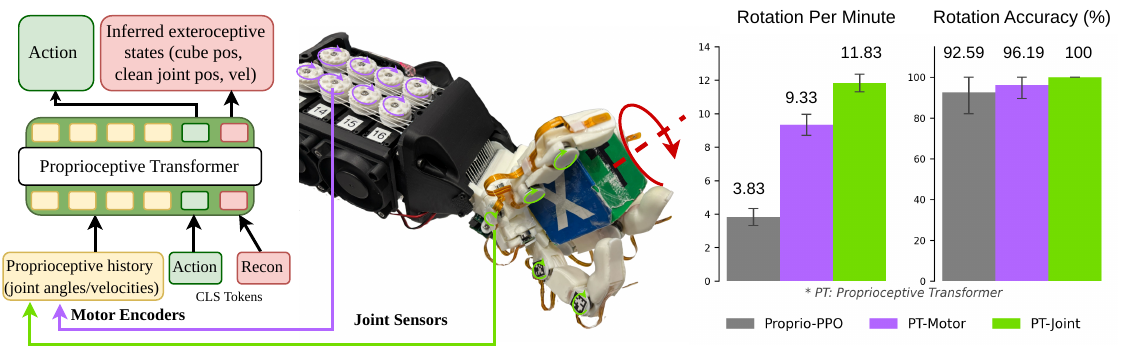}\\[1mm]
    {\small Fig.~1: \textit{Left:} The Proprioceptive Transformer encodes joint position/velocity history and actions, outputting actions and inferred object state. \textit{Middle:} The ORCA hand with 16 AS5600 magnetic angle sensors. \textit{Right:} Our approach achieves $3.1\times$ higher rotation speed and 100\% rotation accuracy. Joint sensors improve rotation speed by 26.8\% over motor encoders.}
    \vspace{-5mm}
\end{strip}
\setcounter{figure}{1}

\begin{abstract}
In-hand object manipulation is a fundamental yet challenging capability for dexterous robots. Despite significant progress in dexterous manipulation, existing approaches rely heavily on vision or tactile sensing to track object states, while joint sensing---the most readily available modality on any robotic hand---remains largely overlooked, particularly for tendon-driven hands. In this paper, we study how far joint sensing alone can go by asking: (i) whether motor encoders or direct joint sensing provides better proprioceptive feedback, (ii) how to extract environment information from joint measurements, and (iii) whether joint-only control can achieve competitive real-world performance without external perception. We present the Proprioceptive Transformer (PT), an exteroceptive-free approach for continuous cube rotation on a tendon-driven dexterous hand that uses only joint sensing feedback. A teacher policy is first trained via reinforcement learning with privileged object information, then distilled into PT, which operates solely on joint position and velocity histories. The Transformer architecture effectively extracts implicit object state information from temporal patterns in joint sensor readings. Experiments on the real ORCA hand show that our approach achieves $3.1\times$ higher rotation speed than baselines. We also demonstrate that our PT achieves a 23.4\% lower RMSE for cube position estimation than the MLP baseline, indicating superior extraction of exteroceptive information from proprioceptive sources.
\end{abstract}

\section{Introduction}

Dexterous manipulation with robotic hands represents one of the most challenging frontiers in robotics research \cite{billard2019trends}. The ability to manipulate objects with human-like dexterity has long been a goal of robotics, yet achieving this capability remains elusive due to high-dimensional action spaces, complex contact dynamics, and the need for precise sensory feedback. Among various manipulation tasks, in-hand object reorientation—the ability to rotate and reposition an object within a robotic hand—has emerged as a canonical benchmark for evaluating dexterous manipulation capabilities \cite{chen2022bidexhands}.

Recent advances in deep reinforcement learning (RL) have shown strong performance in learning complex manipulation policies. OpenAI et al. \cite{openai2020learning} demonstrated that RL agents could learn to reorient a cube to arbitrary target orientations. This achievement relied heavily on large-scale parallel simulation, domain randomization for sim-to-real transfer \cite{tobin2017domain}, and sophisticated vision-based perception systems to track the cube's pose throughout the manipulation process.

However, the reliance on external vision systems introduces several practical limitations. Vision-based approaches require careful camera placement, are susceptible to occlusion during manipulation, and demand substantial computational resources for real-time pose estimation \cite{handa2023dextreme}. Furthermore, the sim-to-real gap for visual perception can be significant, as rendering realistic images in simulation that match real-world conditions remains challenging \cite{zhao2020sim}. These factors motivate the exploration of alternative sensing modalities that could provide robust and efficient feedback for dexterous manipulation tasks.

Proprioceptive sensing---the use of internal sensors to 
perceive the robot's own state---is the most basic and 
widely available sensing modality in robotic hands, yet it 
is still under-explored for dexterous in-hand manipulation~\cite{muratore2022robot}. Joint positions and velocities 
are always available and exhibit a small sim-to-real 
gap. 
This motivates three 
questions.
First, how much exteroceptive information 
can be inferred solely from temporal joint measurements, and 
which policy architecture best captures this information? 
Second, for tendon-driven hands, is it better to 
use motor encoder signals or direct joint sensing as 
proprioceptive feedback, given transmission effects such as 
cable stretch, friction, and backlash? 
Third, can a controller trained from joint 
feedback alone achieve strong real-world in-hand 
manipulation performance without external vision or tactile 
sensing?

To answer these questions, we develop the Proprioceptive 
Transformer (PT) for continuous cube rotation on the ORCA 
tendon-driven hand. PT is trained through a teacher-student 
pipeline: a teacher policy learns with privileged object 
state in simulation, and a student transformer is distilled 
to operate only on joint position and velocity histories. 
Using direct finger-joint magnetic sensors 
\cite{christoph2025orca}, our method avoids motor-side 
transmission mismatch and learns to extract implicit object 
state from proprioceptive sequences. Our key contributions 
are:
\begin{itemize}
    \item A Transformer-based policy that extracts implicit 
    object state from joint sensor history via self-attention, 
    trained with an auxiliary reconstruction objective through 
    teacher-student distillation. We show that this architecture 
    outperforms MLP and LSTM alternatives, and that the 
    reconstruction loss is critical for robust manipulation.
    
    \item Experimental evidence that direct joint sensing via 
    magnetic angle sensors provides superior proprioceptive feedback 
    compared to motor encoder readings, effectively bridging 
    the sim-to-real gap caused by tendon transmission 
    nonlinearities.
    
    \item Real-hardware validation on the ORCA hand 
    demonstrating that joint-only control achieves 
    3.1$\times$ higher rotation speed (11.83 vs. 3.83 RPM) 
    with 100\% accuracy and zero drops, and that joint 
    sensors can implicitly detect object presence and size 
    through position tracking errors.
\end{itemize}

\section{Related Work}

\subsection{Dexterous In-Hand Manipulation}

Dexterous in-hand manipulation has been a longstanding challenge in robotics. Traditional approaches relied on analytical models and trajectory optimization, requiring precise knowledge of object geometry, contact mechanics, and hand kinematics~\cite{bicchi2000hands, okamura2000overview}. However, these model-based methods struggled with the inherent uncertainty of contact dynamics and required extensive manual engineering for each new task.

The advent of deep reinforcement learning (RL) has transformed the field. OpenAI et al.~\cite{openai2020learning} demonstrated that RL agents could learn to reorient a cube to arbitrary target orientations using the Shadow Dexterous Hand. This work leveraged massive parallel simulation and domain randomization to train policies that transferred to real hardware. Building on this success, Akkaya et al.~\cite{akkaya2019solving} showed that similar techniques could solve the Rubik's cube with a robotic hand. Subsequent works have pushed these boundaries further: DeXtreme~\cite{handa2023dextreme} achieved more agile in-hand manipulation with improved sim-to-real transfer, while Bi-DexHands~\cite{chen2022bidexhands} extended these capabilities to bimanual dexterous manipulation.

\subsection{From Vision-Based to Proprioceptive Approaches}

Early successes in cube reorientation relied heavily on vision-based perception systems. OpenAI et al.~\cite{openai2020learning} employed a sophisticated multi-camera setup with pose estimation pipelines to track the cube's position and orientation throughout manipulation. While effective, this approach introduced several practical limitations: vulnerability to occlusion during finger movements, sensitivity to lighting conditions, computational overhead for real-time pose estimation, and a significant sim-to-real gap in visual appearance~\cite{zhao2020sim, tobin2017domain}.

These challenges have motivated a shift toward proprioceptive sensing---using the robot's internal joint sensors rather than external cameras. Proprioceptive observations offer several advantages: they are inherently available in robotic systems, immune to visual occlusions, computationally efficient, and exhibit minimal sim-to-real gap since joint encoder readings are nearly identical in simulation and reality~\cite{muratore2022robot}. Kumar et al.~\cite{kumar2016optimal} demonstrated that complex manipulation behaviors could emerge from proprioceptive feedback combined with tactile sensing. Recent work by Yin et al.~\cite{yin2023rotating} showed that in-hand rotation could be achieved using only proprioceptive and tactile feedback, without any visual input.

\subsection{Joint Sensor Feedback in Dexterous Manipulation}

A critical distinction exists between using motor commands (or motor encoder readings) versus direct joint sensor measurements as proprioceptive feedback. Motor-side measurements can differ significantly from actual joint positions due to transmission effects including gear backlash, cable stretch, and joint compliance~\cite{spong1987modeling, albuschaeffer2007soft}. These discrepancies are particularly problematic for dexterous hands with complex tendon-driven or compliant mechanisms, where the relationship between motor and joint positions is highly nonlinear~\cite{xu2016design}.

Direct joint sensor feedback---measuring positions and velocities at the joint output rather than the motor input---provides more accurate state information. This is especially relevant for hands like the ORCA hand, which features rolling contact joints and tendon-driven actuation~\cite{toshimitsu2023getting, christoph2025orca}. Pitz et al.~\cite{pitz2023dexterous} demonstrated that accurate joint sensing is crucial for learning fine manipulation skills on tendon-driven hands. By using joint encoder readings as policy inputs, we can achieve more precise state estimation and reduce the sim-to-real gap that arises from unmodeled transmission dynamics~\cite{hwangbo2019learning}.

\subsection{Teacher-Student Learning with Transformers}

Teacher-student learning has emerged as a powerful paradigm for robot control, particularly for sim-to-real transfer. In this framework, a teacher policy is first trained in simulation with access to privileged information (such as ground-truth object states), and then a student policy learns to imitate the teacher using only observations available on the real robot~\cite{lee2020learning, chen2020learning}. Kumar et al.~\cite{kumar2021rma} introduced Rapid Motor Adaptation (RMA), which uses a teacher-student framework to enable robust locomotion across diverse terrains.

Transformer architectures have shown strong performance in this context. Unlike recurrent networks, transformers can effectively model long-range temporal dependencies through self-attention mechanisms~\cite{vaswani2017attention}. For robot control, this capability is crucial: the policy can attend to relevant historical observations to infer hidden states such as object pose and contact dynamics. Chen et al.~\cite{chen2021decision} introduced Decision Transformer, framing RL as sequence modeling, while Janner et al.~\cite{janner2021offline} proposed Trajectory Transformer for offline RL. These approaches have been extended to robotics: Brohan et al.~\cite{brohan2022rt} demonstrated RT-1, a transformer-based policy for real-world robot manipulation, and Radosavovic et al.~\cite{radosavovic2024real} achieved impressive humanoid locomotion using transformer policies.

For cube reorientation specifically, transformer-based policies offer an important advantage: they can learn to implicitly estimate the cube's state from sequences of proprioceptive observations, potentially eliminating the need for explicit pose estimation. These advances suggest that combining teacher-student learning with transformer architectures is an effective direction for proprioceptive-based cube manipulation.

\section{Methodology}

\begin{figure}[t]
    \centering

    \includegraphics[width=1.2\linewidth, trim=300 100 100 50, clip]{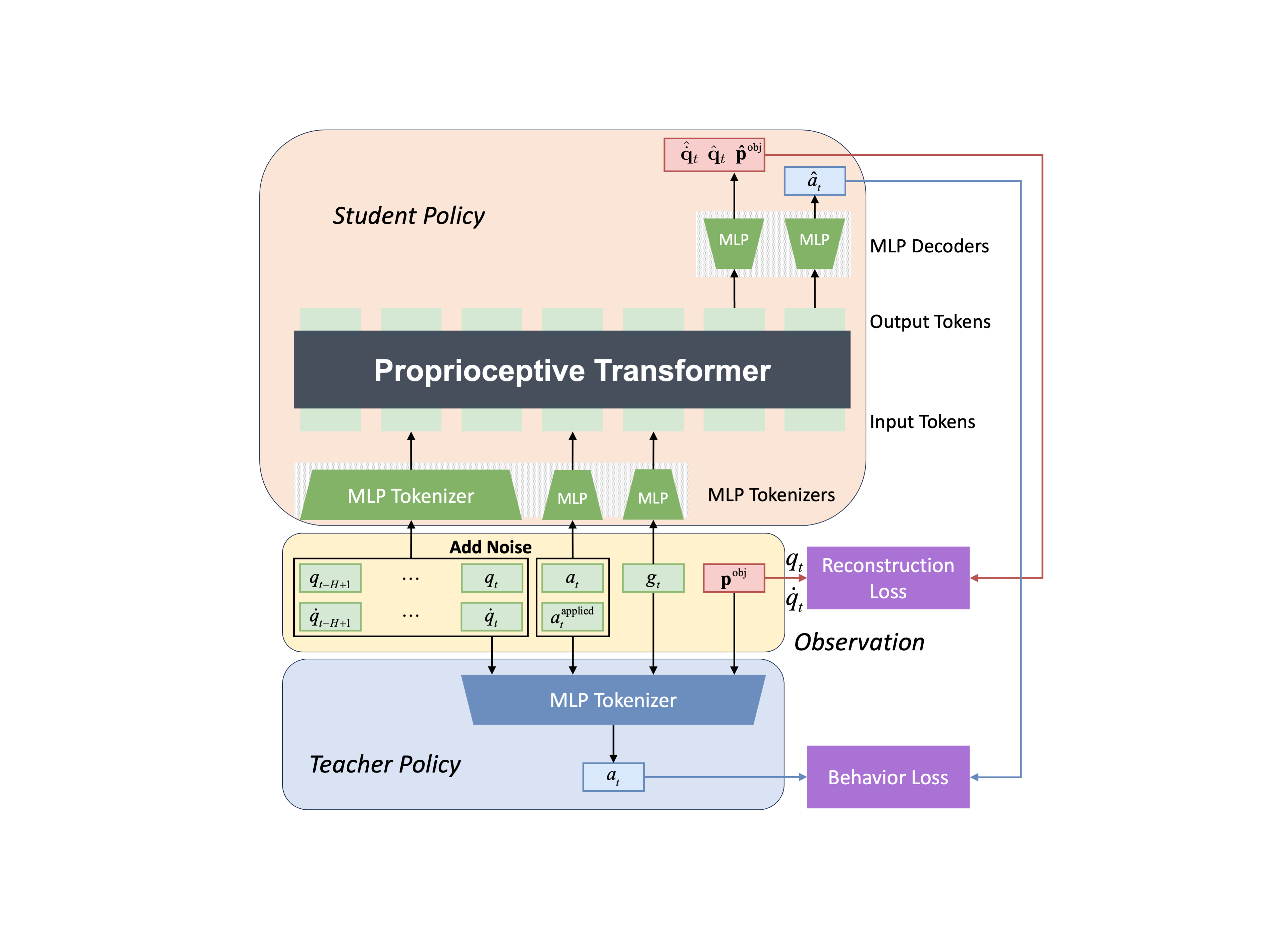}
    \caption{Teacher-student distillation pipeline. The teacher policy is trained via RL with privileged object state in simulation. The student employs a Proprioceptive Transformer encoder to process joint sensor history via self-attention, and is trained to both imitate teacher actions and reconstruct object states and clean joint states from proprioceptive observations alone.}
    \label{fig:pipeline}
\end{figure}

\subsection{Overview}

We propose a three-stage approach for learning cube reorientation using only proprioceptive feedback, as illustrated in Fig.~\ref{fig:pipeline}. First, a teacher policy is trained with reinforcement learning (PPO) using privileged information including ground-truth object pose. Second, a student policy with a proprioceptive transformer architecture is trained to imitate the teacher using only noisy joint sensor readings. Finally, PT is deployed on the real ORCA hand with zero-shot sim-to-real transfer.

\subsection{Training Environment}

We use Isaac Lab as our simulation platform, running 8192 parallel environments of the ORCA~\cite{christoph2025orca} dexterous hand. The simulation operates at 120\,Hz while the policy runs at 20\,Hz (decimation factor of 6), matching the real hardware control frequency. Each episode has a maximum duration of 32 seconds.

\textbf{Task Definition.} The task requires continuous rotation of a 55mm cube following a commanded angular velocity $\boldsymbol{\omega}^* \in \mathbb{R}^3$, constrained to rotation around the z-axis with magnitude up to 1.5\,rad/s. 

\textbf{Domain Randomization.} To facilitate sim-to-real transfer, we randomize physical properties of both the robot and object. For the robot, we scale friction coefficients (0.8--2.0$\times$), link masses (0.95--1.05$\times$), and randomize joint stiffness/damping parameters using log-uniform distributions. For the object, we randomize friction (0.3--1.0), mass (scaled by 0.5--1.5$\times$), and initial pose with full SO(3) rotation sampling.

\subsection{Teacher Policy Training}

The teacher policy is trained using Proximal Policy Optimization (PPO)~\cite{schulman2017proximal} with access to privileged information. We represent the policy as a Gaussian distribution.

\textbf{Observations.} The teacher receives an 81-dimensional noise-free observation vector consisting of: normalized joint positions $\mathbf{q}_t \in \mathbb{R}^{17}$, joint velocities $\dot{\mathbf{q}}_t \in \mathbb{R}^{17}$, ground-truth object position $\mathbf{p}^{\text{obj}}_t \in \mathbb{R}^3$ and quaternion $\mathbf{r}^{\text{obj}}_t \in \mathbb{R}^4$ (privileged), goal command $\mathbf{c}_t \in \mathbb{R}^6$, previous action $\mathbf{a}_{t-1} \in \mathbb{R}^{17}$, and previous joint position command $\mathbf{q}^{\text{cmd}}_{t-1} \in \mathbb{R}^{17}$.

\textbf{Actions.} The policy outputs 17-dimensional joint position targets, mapped from $[-1, 1]$ to actuator limits. To ensure smooth motions, we apply exponential moving average filtering:
\begin{equation}
    a^{\text{applied}}_t = \alpha \cdot a_t + (1 - \alpha) \cdot a^{\text{applied}}_{t-1}
\end{equation}

where $a_t$ is the raw policy output at time step $t$, 
$a_t^{\text{applied}}$ is the smoothed action sent to the 
actuators, and $\alpha$ controls the smoothing strength.

\textbf{Reward Function.} Our reward consists of tracking rewards and regularization terms:
\begin{equation}
    r_t = r_{\text{pos}} + r_{\text{mag}} + r_{\text{dir}} + r_{\text{smooth}} + r_{\text{vel}} + r_{\text{act}} + r_{\text{rate}}
\end{equation}

Position tracking uses an exponential kernel:
\begin{equation}
    r_{\text{pos}} = w_p \cdot \exp\left(-\frac{\|\mathbf{p}^{\text{obj}} - \mathbf{p}^{\text{obj}*}\|}{\sigma_p}\right)
\end{equation}
where $\mathbf{p}^{\text{obj}}$ and 
$\mathbf{p}^{\text{obj}*}$ are the current and target 
object positions, $w_p$ is the reward weight, and 
$\sigma_p$ is the temperature parameter.

Angular velocity tracking is decomposed into magnitude and direction:
\begin{equation}
    r_{\text{mag}} = w_m \cdot \exp\left(-\frac{|\|\boldsymbol{\omega}\| - \|\boldsymbol{\omega}^*\||}{\sigma_\omega}\right)
\end{equation}
\begin{equation}
    r_{\text{dir}} = w_d \cdot \cos(\boldsymbol{\omega}, \boldsymbol{\omega}^*)
\end{equation}
where $\boldsymbol{\omega}$ and 
$\boldsymbol{\omega}^*$ are the current and target 
angular velocities, $w_m$ and $w_d$ are reward weights, 
and $\sigma_\omega$ is the temperature parameter.

Regularization terms penalize angular acceleration ($r_{\text{smooth}}$), joint velocities ($r_{\text{vel}}$), action magnitude ($r_{\text{act}}$), and action rate ($r_{\text{rate}}$). The action rate penalty proved critical for producing smooth motions that transfer to hardware. All reward weights are reported in Section~\ref{sec:setup}.

\subsection{Student Policy Training}

\textbf{Student Policy Observation.} The student operates without privileged information, receiving only: (1) noisy joint positions and velocities over a configurable history window of $T$ steps, (2) previous action and joint position command, and (3) goal command. Joint positions are corrupted with two independent 
Gaussian noise terms: a per-episode bias 
$b_j \sim \mathcal{N}(0, \sigma_b^2)$ sampled once at 
episode start, and a per-step noise 
$n_t \sim \mathcal{N}(0, \sigma_n^2)$ sampled at each 
time step. Joint velocities are corrupted with 
per-step Gaussian noise.

\textbf{Network Architecture.} As shown in Fig.~\ref{fig:pipeline}, each observation group is first projected to a 256-dimensional token embedding via a linear layer, producing 10 tokens from the proprioceptive history and 1 token each from the action context and command groups (12 tokens total). Sinusoidal positional embeddings are added across the full sequence. A Transformer encoder (3 layers, 4 heads, feedforward dim 1024) processes the sequence with learnable \texttt{[CLS]} query tokens appended, which are mapped through separate linear heads to predict actions $\hat{\mathbf{a}}_t \in \mathbb{R}^{17}$ and auxiliary reconstructions of object position and clean joint states.

\textbf{Training Losses.} The behavior cloning loss 
minimizes the L2 distance between student and teacher 
actions:
\begin{equation}
    \mathcal{L}_{\text{BC}} = \|\hat{\mathbf{a}}_t - 
    \mathbf{a}^{\text{teacher}}_t\|_2^2
\end{equation}
where $\hat{\mathbf{a}}_t$ is the student's predicted 
action and $\mathbf{a}^{\text{teacher}}_t$ is the 
teacher's action at time step $t$.

The auxiliary reconstruction loss encourages the Transformer to encode implicit object state and denoise joint measurements from proprioceptive history:
\begin{equation}
    \mathcal{L}_{\text{recon}} = \lambda_p \|\hat{\mathbf{p}}^{\text{obj}}_t - \mathbf{p}^{\text{obj}}_t\|_2^2 + \lambda_q \|\hat{\mathbf{q}}_t - \mathbf{q}_t\|_2^2 + \lambda_v \|\hat{\dot{\mathbf{q}}}_t - \dot{\mathbf{q}}_t\|_2^2
\end{equation}
where $\hat{\mathbf{p}}^{\text{obj}}_t$ is the predicted object position, $\hat{\mathbf{q}}_t$ and $\hat{\dot{\mathbf{q}}}_t$ are the predicted clean joint positions and velocities, and unhatted quantities denote ground-truth values available only in simulation. $\lambda_p$, $\lambda_q$, $\lambda_v$ weight each term respectively. The total loss is $\mathcal{L} = \mathcal{L}_{\text{BC}} + \mathcal{L}_{\text{recon}}$.

\section{Experiments}
\label{sec:setup}

\begin{figure*}[t]
    \centering
    \includegraphics[width=\linewidth]{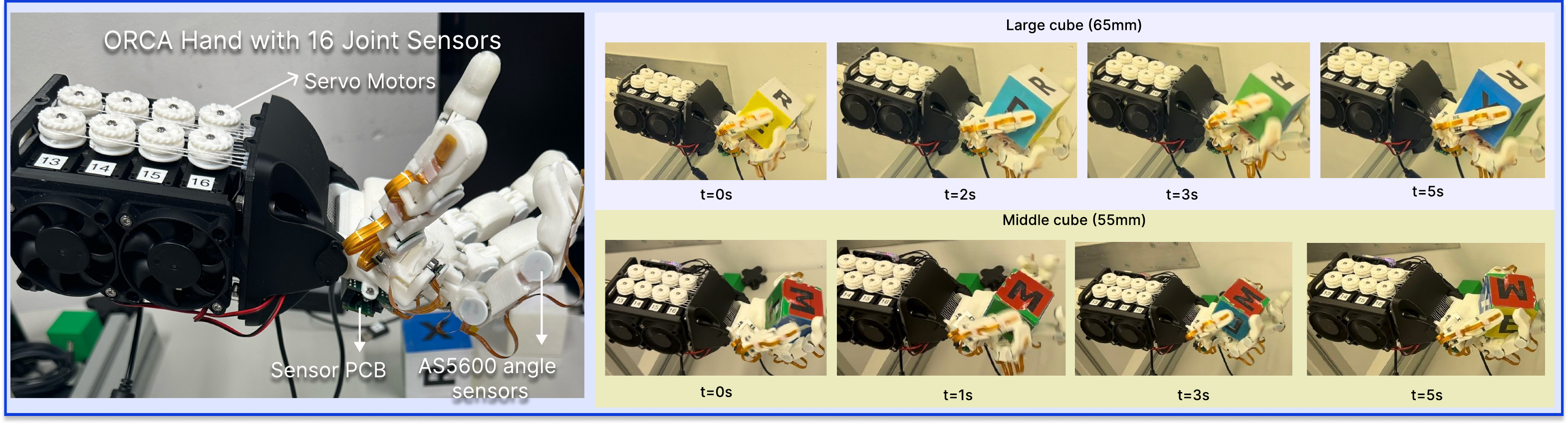}
    \caption{\textit{Left:} The ORCA hand in our real-world experiment. Magnetic angle sensors are embedded in the joints for direct joint sensing. The antagonistic tendon design allows the joint angles to be estimated from motor angle readings. \textit{Right:} Continuous cube rotation on 55\,mm and 65\,mm cubes with our PT-Joint Policy.}
    \label{fig:realworld}
\end{figure*}

\subsection{Experimental Setup}

We evaluate our approach with the ORCA hand
~\cite{christoph2025orca}, a tendon-driven dexterous 
manipulator with 17 actuated degrees of freedom (16 
finger joints and 1 wrist joint) powered by Dynamixel 
servo motors, as shown in Fig.~\ref{fig:realworld}. The 
hand is mounted in a palm-up configuration. For 
proprioceptive sensing, 16 magnetic angle sensors 
(AS5600, ams OSRAM AG) are integrated at the finger 
joints, measuring joint angles directly independent of 
tendon stretch or motor backlash. The 16 AS5600 sensors are mounted at the finger joints, so the wrist joint angle is not directly measured. During deployment, the wrist position is fixed at its zero configuration. Joint velocities are 
computed by differentiating sensor readings. The control 
system runs at 20\,Hz via ROS2. The trained policy is 
deployed with direct sim-to-real transfer, taking joint 
positions and velocities as input and outputting joint 
position targets to the low-level PD controller. We 
evaluate on two cube sizes: a 55\,mm medium cube and a 
65\,mm large cube, to test generalization across object 
dimensions.

\textbf{Implementation Details.} All training is 
conducted in Isaac Lab with 8192 parallel environments 
on a single NVIDIA RTX 4070 Ti SUPER GPU.

The teacher policy uses a four-layer MLP with hidden 
dimensions [512, 512, 256, 128] and ELU activations, 
trained for 10,000 PPO iterations with learning rate 
$10^{-4}$. The tracking reward weights are position 
($w_p = 10$, $\sigma_p = 0.02$), angular velocity 
magnitude ($w_m = 30$), and direction ($w_d = 60$). 
Regularization penalties include angular acceleration 
($w_{\text{smooth}} = 1$), joint velocity 
($w_{\text{vel}} = -10^{-5}$), action magnitude 
($w_{\text{act}} = -2 \times 10^{-4}$), and action rate 
($w_{\text{rate}} = -0.075$).

The student policy (deployed as PT) is trained for 1200 epochs (6000 gradient update steps) with batch size 256, learning rate $10^{-5}$, weight decay 0.01, and reconstruction loss weights $\lambda_p = 0.5$, $\lambda_q = 0.3$, $\lambda_v = 0.2$.

\textbf{Evaluation Metrics.} We introduce four metrics: (1) \textit{Rotations Per Minute (RPM)}: complete rotations around the target axis per minute; (2) \textit{Rotation Accuracy (RA)}: ratio of correct-direction rotations to total z-axis rotations; (3) \textit{Drop-Free Success Rate (DFSR)}: ratio of rotations completed without dropping; (4) \textit{Drop Count (DC)}: total drops per trial. Each trial lasts 60 seconds and results are averaged over 3 trials.

\textbf{Baselines.} We denote our method as \textbf{Proprioceptive Transformer (PT)}, with two variants: 
\textit{PT-Joint}, which uses direct joint sensor 
feedback via AS5600 magnetic angle sensors, and 
\textit{PT-Motor}, which relies on motor encoder 
readings only. We compare against: (1) 
\textit{Proprio-PPO}: PPO policy using only joint 
positions and velocities without object information; 
(2) \textit{Extero-PPO}: PPO policy trained with 
ground-truth object pose in simulation. During real-world deployment, the object pose is detected from camera images similar to~\cite{handa2023dextreme}.

\subsection{Proprioceptive Transformer vs. PPO Baselines}

\begin{table}[t]
    \centering
    \caption{Performance comparison on cube rotation task. Best results in \textbf{bold}.}
    \label{tab:main}
    \setlength{\tabcolsep}{4pt}
    \renewcommand{\arraystretch}{1.1}
    \begin{tabular}{l|cccc}
    \hline
    \multicolumn{5}{c}{\textbf{Medium Cube (55\,mm)}} \\
    \hline
    Policy & RPM$\uparrow$ & RA(\%)$\uparrow$ & DFSR(\%)$\uparrow$ & DC$\downarrow$ \\
    \hline
    Proprio-PPO & 3.83\tiny{$\pm$0.51} & 92.59\tiny{$\pm$10.47} & 89.74\tiny{$\pm$14.51} & 0.33\tiny{$\pm$0.58} \\
    Extero-PPO & 3.08\tiny{$\pm$0.12} & \textbf{100}\tiny{$\pm$0} & \textbf{100}\tiny{$\pm$0} & \textbf{0} \\
    \hline
    \multicolumn{5}{l}{\textit{Ours}} \\
    PT-Motor & 9.33\tiny{$\pm$0.63} & 
    96.19\tiny{$\pm$6.60} & 98.10\tiny{$\pm$3.30} & 0.33\tiny{$\pm$0.58} \\
    PT-Joint & \textbf{11.83}\tiny{$\pm$0.52} & \textbf{100}\tiny{$\pm$0} & \textbf{100}\tiny{$\pm$0} & \textbf{0} \\
    \hline
    \multicolumn{5}{c}{\textbf{Large Cube (65\,mm)}} \\
    \hline
    Policy & RPM$\uparrow$ & RA(\%)$\uparrow$ & DFSR(\%)$\uparrow$ & DC$\downarrow$ \\
    \hline
    Proprio-PPO & 5.17\tiny{$\pm$0.12} & 86.98\tiny{$\pm$9.21} & \textbf{100}\tiny{$\pm$0} & \textbf{0} \\
    Extero-PPO & 4.83\tiny{$\pm$0.12} & \textbf{100}\tiny{$\pm$0} & \textbf{100}\tiny{$\pm$0} & \textbf{0} \\
    \hline
    \multicolumn{5}{l}{\textit{Ours}} \\
    PT-Motor & 8.53\tiny{$\pm$1.48} & \textbf{100}\tiny{$\pm$0} & 96.34\tiny{$\pm$5.18} & 0.33\tiny{$\pm$0.58}  \\
    PT-Joint & \textbf{11.33}\tiny{$\pm$0.12} & \textbf{100}\tiny{$\pm$0} & \textbf{100}\tiny{$\pm$0} & \textbf{0} \\
    \hline
    \end{tabular}
    \vspace{-2mm}
\end{table}

Each policy is evaluated over 3 trials of 60 seconds 
each. Table~\ref{tab:main} presents the quantitative comparison 
on the real robot. Both PT variants significantly 
outperform all baselines across both cube sizes.

Compared to Proprio-PPO, PT-Joint achieves 
3.1$\times$ improvement in RPM (11.83 vs. 3.83 
for medium cube) while maintaining perfect rotation accuracy 
and zero drops. Notably, Proprio-PPO exhibits considerable 
variance in RA (92.59$\pm$10.47\%) and DFSR 
(89.74$\pm$14.51\%), whereas both PT variants achieve 
consistent 100\% accuracy with zero variance, indicating 
superior robustness to observation noise and domain shift.
These advantages generalize to the large cube, where 
PT-Joint achieves 11.33 RPM---2.2$\times$ and 
2.3$\times$ higher than Proprio-PPO and Extero-PPO, 
respectively---demonstrating robust transfer across 
object sizes without retraining.

PT-Joint also surpasses Extero-PPO despite the latter 
having access to ground-truth object 
poses---achieving 3.8$\times$ higher RPM (11.83 vs. 3.08) 
for the medium cube. This suggests that implicit state 
estimation from proprioceptive history can be more 
effective than explicit but noisy pose feedback in 
sim-to-real transfer.

Comparing PT-Joint and PT-Motor, direct joint sensing 
provides a consistent 26.8\% RPM improvement (11.83 vs. 
9.33 for medium cube), confirming that joint sensors better 
capture the true kinematic state by bypassing tendon 
transmission nonlinearities. Table~\ref{tab:ablation} 
further examines this comparison across different 
observation window sizes, showing that PT-Joint maintains 
100\% RA and DFSR at all window sizes while PT-Motor 
degrades significantly at smaller windows.

\subsection{Implicit Object Detection via Joint Sensors}
\label{sec:joint_sensor}
\begin{figure*}[t]
    \centering
    \includegraphics[width=\textwidth]{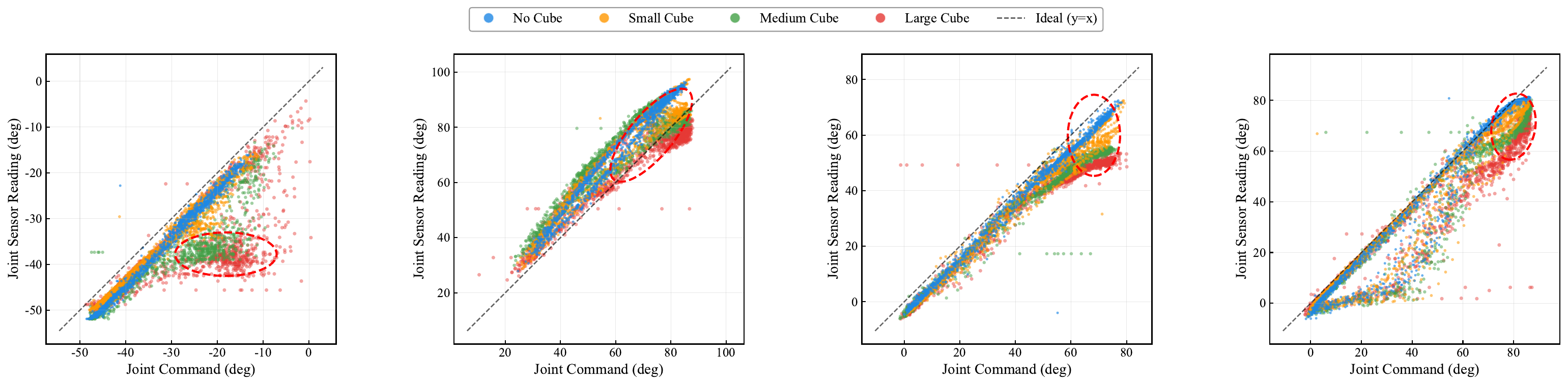}
    \caption{Joint command vs.\ actual position for different cube sizes. 
    Deviation from the ideal line ($y{=}x$) increases with cube size due to 
    volume blocking and weight-induced deformation. 
    (a)~Thumb MCP. (b)~Thumb PIP. (c)~Index MCP. (d)~Ring MCP.}
    \label{fig:joint_analysis}
\end{figure*}

To better understand how the Proprioceptive Transformer extracts exteroceptive information from proprioceptive observations---and whether direct joint sensing provides a more informative signal than motor encoders---we analyze how joint sensors implicitly encode object information by examining the relationship between joint commands and actual joint positions under different object conditions. Fig.~\ref{fig:joint_analysis} shows scatter plots for representative 
joints across four conditions collected in simulation: no cube, 
45\,mm small cube, 55\,mm medium cube, and 65\,mm large cube.

We observe two distinct effects as cube size increases:

\textbf{Volume Effect.} Larger cubes physically block fingers from reaching commanded positions. When a finger encounters the cube surface, it cannot complete the commanded motion, resulting in position errors. This is particularly visible in MCP joints, where different object volumes lead to distinct grasp postures, as highlighted by the circled regions in Fig.~\ref{fig:joint_analysis}(c) and (d).

\textbf{Weight Effect.} Heavier cubes exert more downward force on the fingers. Under the same PD controller gains, this increased load leads to larger steady-state position errors, particularly in the thumb joints that support the object from below, as shown by the circled region in Fig.~\ref{fig:joint_analysis}(a) and (b).

These effects create characteristic signatures in joint position data that correlate with object size and presence. Without any cube, actual joint positions closely follow commands (data points near $y{=}x$ line). As cube size increases, deviations become progressively larger, forming distinguishable patterns across object conditions, suggesting that joint sensors can implicitly encode object presence and size through position tracking errors---without vision or tactile feedback.

\subsection{Implicit State Estimation via Transformer}
\label{sec:implicit_state}

\begin{figure*}[t]
\centering
\begin{subfigure}[b]{0.45\textwidth}
    \includegraphics[width=\textwidth]{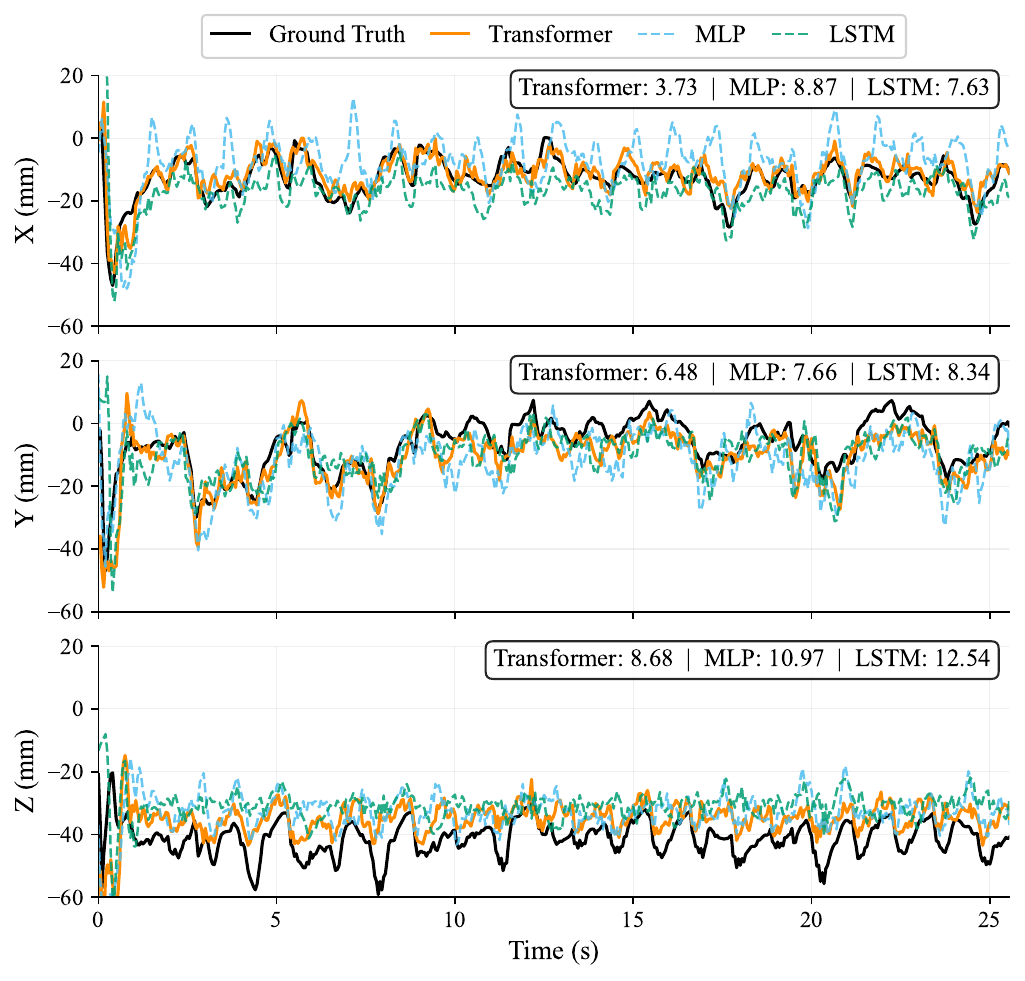}
    \caption{}
    \label{fig:recon_xyz}
\end{subfigure}
\hspace{0.02\textwidth}
\begin{subfigure}[b]{0.45\textwidth}
    \includegraphics[width=\textwidth]{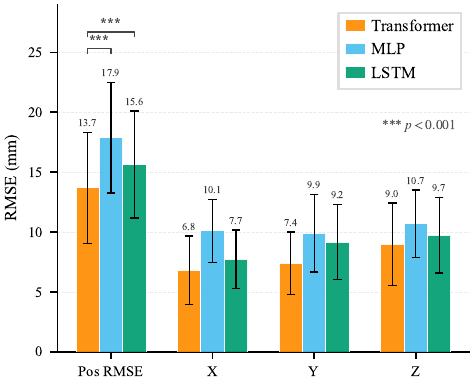}
    \caption{}
    \label{fig:recon_bar}
\end{subfigure}
\caption{Object position reconstruction from proprioceptive observations alone.
(a)~Per-axis tracking for a representative environment.
(b)~Reconstruction RMSE across 32 environments (mean~$\pm$~std); the Transformer achieves 13.7\,mm, outperforming MLP (17.9\,mm, $-$23.4\%) and LSTM (15.6\,mm, $-$12.2\%).
*** denotes $p < 0.001$ (paired $t$-test with Bonferroni correction).}
\label{fig:recon_multi}
\end{figure*}

To assess whether temporal modeling is critical for reconstruction quality, we compare three encoder architectures: a Transformer (3-layer, 4-head), an MLP (equivalent parameter count), and an LSTM. All variants share the same reconstruction head and training procedure, differing only in how the observation history is processed. We evaluate reconstruction on both object position and joint state (position and velocity), as accurate joint state recovery from noisy sensor readings is a prerequisite for reliable sim-to-real transfer.

\textbf{Object State Reconstruction.} Fig.~\ref{fig:recon_xyz} shows a representative reconstruction trajectory for environment~21, where the cube undergoes complex rotation dynamics. The Transformer (orange) closely follows the ground-truth position (black) across all three axes. Notably, during rapid position transitions---such as the sharp drop along Y near $t = 1$\,s caused by the cube shifting under finger contact, and the high-frequency oscillations along X between $t = 5$--$10$\,s corresponding to active reorientation phases---the Transformer captures these transient dynamics with minimal delay. In contrast, the MLP (blue) exhibits systematic offset, particularly along the Y-axis where it fails to track the cube's lateral displacement during rotation. The LSTM (green) shows improved temporal sensitivity over MLP but still diverges during rapid state changes, especially along Z where gravity-coupled dynamics require full-sequence attention to model accurately.

Fig.~\ref{fig:recon_bar} quantifies the reconstruction accuracy across all 32 simulation environments. The Transformer achieves the lowest position RMSE of $13.70 \pm 4.62$\,mm, compared to $17.87 \pm 4.60$\,mm for MLP and $15.64 \pm 4.46$\,mm for LSTM. The Transformer consistently outperforms both baselines across all individual axes, with the largest improvement observed along the X-axis ($6.82$ vs.\ $10.11$\,mm for MLP, a 33\% reduction). Pairwise paired $t$-tests with Bonferroni correction confirm that all differences are statistically significant: Transformer vs.\ MLP ($p < 0.001$, Cohen's $d = 1.35$) and Transformer vs.\ LSTM ($p < 0.001$, $d = 1.12$), both with large effect sizes. Notably, the Z-axis exhibits the largest error across all architectures due to gravity-induced finger deflection, yet even this worst case remains well below the cube dimension (55\,mm).

\begin{figure}[t]
    \centering
    \includegraphics[width=\columnwidth]{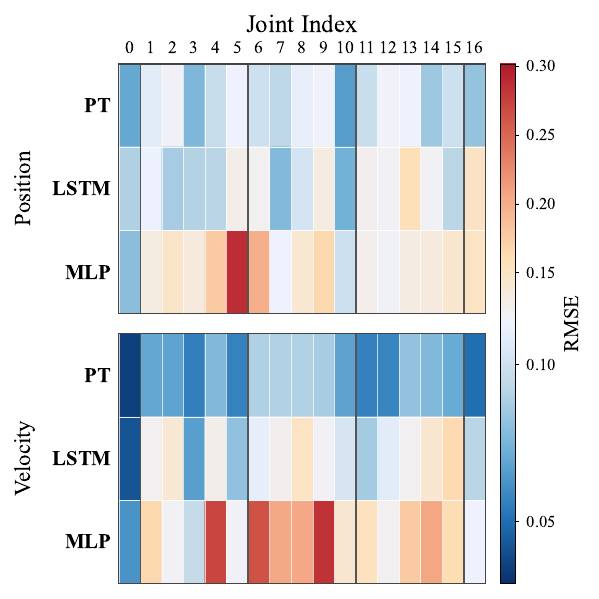}
    \caption{Per-joint reconstruction RMSE across architectures. Joints are grouped by type: 0\,=\,Wrist; 1--5\,=\,Abduction (Index, Middle, Pinky, Ring, Thumb); 6--10\,=\,Metacarpophalangeal; 11--15\,=\,Proximal interphalangeal; 16\,=\,Distal interphalangeal (Thumb). Each group follows the same finger order.}
    \label{fig:joint_heatmap}
\end{figure}

\textbf{Joint State Reconstruction.} Beyond object state, we evaluate each architecture's ability to reconstruct noise-free joint positions and velocities from noisy proprioceptive observations. Fig.~\ref{fig:joint_heatmap} presents the per-joint RMSE grouped by joint type. The Transformer achieves the lowest mean position RMSE ($0.098$\,rad) and velocity RMSE ($0.070$\,rad/s), outperforming LSTM ($0.110$\,rad, $0.112$\,rad/s) and MLP ($0.146$\,rad, $0.168$\,rad/s). MCP joints exhibit the highest error across all architectures, reflecting complex contact dynamics at these joints. The MLP fails most on abduction joints (e.g., Thumb ABD: $0.287$ vs.\ $0.114$\,rad for Transformer), confirming that temporal context is necessary to disentangle contact-induced deflections from voluntary motion. These results show the Transformer can effectively denoise joint measurements, which is critical for robust sim-to-real transfer.

\subsection{Ablation Study}
\label{sec:ablation}
All ablation experiments are conducted on the real robot using the 55\,mm medium cube. We conduct ablation studies on observation window size, sensor feedback configuration, encoder architecture, and reconstruction objective. Tables~\ref{tab:ablation} and~\ref{tab:encoder_ablation} summarize the results. Each configuration is evaluated over 3 trials of 
60 seconds each.

\begin{table}[t]
    \centering
    \caption{Effect of observation window size and sensor feedback.}
    \label{tab:ablation}

    \textbf{(a) w/ Sensor (PT-Joint)}\\[2pt]
    \begin{tabular*}{\columnwidth}{@{\extracolsep{\fill}}c|cccc}
        \toprule
        Win. & RPM & RA(\%) & DFSR(\%) & DC \\
        \midrule
        1  & $8.33_{\pm0.14}$  & $100_{\pm0}$ & $100_{\pm0}$ & $0_{\pm0}$ \\
        3  & $8.17_{\pm1.04}$  & $100_{\pm0}$ & $100_{\pm0}$ & $0_{\pm0}$ \\
        6  & $8.75_{\pm0.25}$  & $100_{\pm0}$ & $100_{\pm0}$ & $0_{\pm0}$ \\
        10 & $\mathbf{11.83}_{\pm0.52}$ & $100_{\pm0}$ & $100_{\pm0}$ & $0_{\pm0}$ \\
        \midrule
        \textit{Mean} & \textit{9.27} & \textit{100} & \textit{100} & \textit{0} \\
        \bottomrule
    \end{tabular*}

    \vspace{4pt}
    \textbf{(b) w/o Sensor (PT-Motor)}\\[2pt]
    \begin{tabular*}{\columnwidth}{@{\extracolsep{\fill}}c|cccc}
        \toprule
        Win. & RPM & RA(\%) & DFSR(\%) & DC \\
        \midrule
        1  & $7.83_{\pm0.14}$ & $97.0_{\pm5.2}$   & $100_{\pm0}$      & $0_{\pm0}$ \\
        3  & $9.25_{\pm1.09}$ & $73.9_{\pm14.3}$  & $78.2_{\pm10.3}$  & $2_{\pm1}$ \\
        6  & $8.33_{\pm0.14}$ & $71.9_{\pm7.2}$   & $100_{\pm0}$      & $0_{\pm0}$ \\
        10 & $9.33_{\pm0.63}$ & $96.19_{\pm6.60}$ & $98.10_{\pm3.30}$ & $0.33_{\pm0.58}$ \\
        \midrule
        \textit{Mean} & \textit{8.69} & \textit{84.75} & \textit{94.08} & \textit{0.58} \\
        \bottomrule
    \end{tabular*}
\end{table}

\textbf{Effect of Sensor Feedback.} Across all window sizes, configurations with joint sensors achieve perfect consistency: 100\% RA and 100\% DFSR. Without sensors, RA drops to 84.75\% and DFSR to 94.08\% on average. The PT-Joint policy also achieves 6.7\% higher RPM (9.27 vs. 8.69) than PT-Motor policy. Direct joint measurements from joint sensors provide reliable feedback that effectively bridges the sim-to-real gap.

\textbf{Effect of Window Size.} Larger temporal context improves performance. As the window size increases from 1 to 10, the average RPM improves by 31\% (8.08 to 10.58, averaged across sensor configurations). A window size of 10 achieves best results across all metrics, demonstrating that the transformer encoder effectively leverages observation history to implicitly estimate object state through self-attention.

\begin{table}[t]
    \centering
    \caption{Effect of encoder architecture and reconstruction loss.}
    \label{tab:encoder_ablation}
    \begin{tabular}{l|cccc}
        \toprule
        Policy & RPM$\uparrow$ & RA(\%)$\uparrow$ & DFSR(\%)$\uparrow$ & DC$\downarrow$ \\
        \midrule
        No Recon    & $4.75_{\pm0.75}$  & $68.9_{\pm39.1}$  & $35.0_{\pm20.9}$  & $5.0_{\pm1.7}$ \\
        MLP         & $5.08_{\pm0.38}$  & $68.1_{\pm19.7}$  & $61.5_{\pm16.8}$  & $2.0_{\pm1.0}$ \\
        LSTM        & $7.00_{\pm0.25}$  & $95.4_{\pm8.0}$   & $100_{\pm0}$      & $0.0_{\pm0.0}$ \\
        Transformer & $9.33_{\pm0.63}$ & $96.19_{\pm6.60}$ & $98.10_{\pm3.30}$ & $0.33_{\pm0.58}$ \\
        \bottomrule
    \end{tabular}
\end{table}
\textbf{Effect of Encoder Architecture and Reconstruction.} We evaluate encoder architecture and reconstruction objective under the w/o Sensor configuration (window size 10) to isolate architectural effects from sensor type. The Transformer achieves the highest RPM ($9.33_{\pm0.63}$) with 96.19\% RA and 98.10\% DFSR, outperforming LSTM ($7.00_{\pm0.25}$ RPM) and MLP ($5.08_{\pm0.38}$ RPM) by 33\% and 84\%, respectively. This demonstrates that self-attention more effectively extracts implicit object state from proprioceptive history than recurrent or feedforward alternatives, by selectively attending to contact-relevant time steps. Ablating the reconstruction head (No Recon) causes severe degradation—DFSR drops to $35.0_{\pm20.9}$\% with an average of 5 drops per trial—confirming that the auxiliary reconstruction objective is critical for shaping latent representations that capture object dynamics.

\section{Conclusion}

Our work demonstrates that high-performance in-hand manipulation can be achieved from proprioception alone.
We introduced the Proprioceptive Transformer, a teacher-student framework that distills privileged simulation information into a deployable policy using only joint position and velocity histories, and we showed that direct joint sensing is more effective than motor-side sensing for tendon-driven hands.

On real-world benchmarks with the ORCA hand, PT-Joint achieved a \textbf{3.1$\times$} speed gain over the baseline on the medium cube and up to \textbf{2.3$\times$} over Extero-PPO on the large cube. Ablation and reconstruction analyses further support these results: removing the reconstruction objective reduced DFSR to \textbf{35.0\%} with an average of 5 drops, and the PT achieved the lowest object-state reconstruction error 23.4\% lower than the MLP baseline.

More broadly, these results support a practical direction for dexterous manipulation: extracting exteroceptive information from onboard proprioception to reduce dependence on external perception stacks. Future work will extend this framework to multi-axis reorientation and more diverse object geometries, and will study how proprioceptive latent-state learning can be combined with sparse exteroceptive cues (such as tactile information) for even broader real-world generalization.






\bibliographystyle{IEEEtran}
\bibliography{references}
\end{document}